\documentclass{article}
\usepackage[utf8]{inputenc}
\usepackage{main}
\pdfoutput=1
\usepackage{microtype}
\usepackage{graphicx}
\usepackage{subfig}
\usepackage{times}
\usepackage{latexsym}
\usepackage{amsmath}
\usepackage{float}
\usepackage{footnote}
\usepackage{enumitem}
\usepackage{bm}
\usepackage{arydshln}
\usepackage{booktabs}
\usepackage{multicol}
\usepackage{multirow}
\usepackage{color}
\usepackage{xcolor}     
\usepackage{colortbl}
\usepackage{bbding}
\usepackage{makecell}
\usepackage{mathtools}
\usepackage{imakeidx}
\usepackage{longtable}
\usepackage{wrapfig}
\usepackage{picinpar}
\makeindex
\usepackage{arydshln}
\usepackage{lipsum}
\usepackage{natbib}
\usepackage[toc]{multitoc}
\usepackage[edges]{forest}
\usepackage[normalem]{ulem}
\definecolor{mydarkblue}{rgb}{0,0.08,0.45}
\usepackage[colorlinks=true,linkcolor=mydarkblue,citecolor=mydarkblue,filecolor=mydarkblue,urlcolor=mydarkblue]{hyperref}
\usepackage{caption}
\usepackage{CJKutf8}
\usepackage{awesomebox} 
\usepackage{bbding}
\usepackage[most]{tcolorbox}
\usepackage{amssymb}
\usepackage{enumitem}

\usepackage{hyperref}

\usepackage[tikz]{bclogo}
\usepackage[framemethod=tikz]{mdframed}
\definecolor{bgblue}{RGB}{245,243,253}
\definecolor{ttblue}{RGB}{91,194,224}

\mdfdefinestyle{mystyle}{%
  rightline=true,
  innerleftmargin=10,
  innerrightmargin=10,
  outerlinewidth=3pt,
  topline=false,
  rightline=true,
  bottomline=false,
  skipabove=\topsep,
  skipbelow=\topsep
}

\newtcolorbox{myboxi}[1][]{
  breakable,
  title=#1,
  colback=red!5,
  colbacktitle=red!5,
  coltitle=black,
  fonttitle=\bfseries,
  bottomrule=0pt,
  toprule=0pt,
  leftrule=2pt,
  rightrule=2pt,
  titlerule=0pt,
  arc=0pt,
  outer arc=0pt,
  colframe=red,
}

\newtcolorbox{myboxnote}[1][]{
  breakable,
  title=#1,
  colback=orange!0,
  colbacktitle=orange!0,
  coltitle=black,
  fonttitle=\bfseries,
  bottomrule=0pt,
  toprule=0pt,
  leftrule=2pt,
  rightrule=2pt,
  titlerule=0pt,
  arc=0pt,
  outer arc=0pt,
  colframe=orange,
}

\newtcolorbox{myboxii}[1][]{
  breakable,
  freelance,
  title=#1,
  colback=white,
  colbacktitle=white,
  coltitle=black,
  fonttitle=\bfseries,
  bottomrule=0pt,
  boxrule=0pt,
  colframe=white,
  overlay unbroken and first={
  \draw[red!75!black,line width=3pt]
    ([xshift=5pt]frame.north west) -- 
    (frame.north west) -- 
    (frame.south west);
  \draw[red!75!black,line width=3pt]
    ([xshift=-5pt]frame.north east) -- 
    (frame.north east) -- 
    (frame.south east);
  },
  overlay unbroken app={
  \draw[red!75!black,line width=3pt,line cap=rect]
    (frame.south west) -- 
    ([xshift=5pt]frame.south west);
  \draw[red!75!black,line width=3pt,line cap=rect]
    (frame.south east) -- 
    ([xshift=-5pt]frame.south east);
  },
  overlay middle and last={
  \draw[red!75!black,line width=3pt]
    (frame.north west) -- 
    (frame.south west);
  \draw[red!75!black,line width=3pt]
    (frame.north east) -- 
    (frame.south east);
  },
  overlay last app={
  \draw[red!75!black,line width=3pt,line cap=rect]
    (frame.south west) --
    ([xshift=5pt]frame.south west);
  \draw[red!75!black,line width=3pt,line cap=rect]
    (frame.south east) --
    ([xshift=-5pt]frame.south east);
  },
}

\usepackage{fancyhdr} 
\usepackage{blindtext} 

\DeclareCaptionFont{black}{\color{black}}

\definecolor{myblue}{rgb}{0.9, 0.1, 0.94}
\definecolor{mygreen}{rgb}{0.64, 0.56, 0.88}
\definecolor{myyellow}{rgb}{0.68, 0.6, 0.1}
\definecolor{fancygreen}{rgb}{0.33, 0.68, 0.20}
\definecolor{salmon}{rgb}{0.94, 0.52, 0.49}
\definecolor{tablegreen}{rgb}{0.82, 0.94, 0.75}
\definecolor{tableblue}{rgb}{0.81, 0.90, 0.94}
\definecolor{tablered}{rgb}{0.97, 0.85, 0.85}
\definecolor{tableorange}{rgb}{0.96, 0.85, 0.81}

\newenvironment{itemize*}%
 {\leftmargini=10pt\begin{itemize}%
  \setlength{\itemsep}{0pt}%
  \setlength{\parskip}{0pt}%
  }%
 {\end{itemize}}
\newenvironment{enumerate*}%
 {\begin{enumerate}%
  \setlength{\itemsep}{0pt}%
  \setlength{\parskip}{0pt}}%
 {\end{enumerate}}

\usepackage{xcolor}
\usepackage{listings}

\newcommand\JSONnumbervaluestyle{\color{blue}}
\newcommand\JSONstringvaluestyle{\color{red}}

\newif\ifcolonfoundonthisline

\makeatletter

\lstdefinestyle{json}
{
  showstringspaces    = false,
  keywords            = {false,true},
  alsoletter          = 0123456789.,
  morestring          = [s]{"}{"},
  stringstyle         = \ifcolonfoundonthisline\JSONstringvaluestyle\fi,
  MoreSelectCharTable =%
    \lst@DefSaveDef{`:}\colon@json{\processColon@json},
  basicstyle          = \ttfamily,
  keywordstyle        = \ttfamily\bfseries,
}

\newcommand\processColon@json{%
  \colon@json%
  \ifnum\lst@mode=\lst@Pmode%
    \global\colonfoundonthislinetrue%
  \fi
}

\lst@AddToHook{Output}{%
  \ifcolonfoundonthisline%
    \ifnum\lst@mode=\lst@Pmode%
      \def\lst@thestyle{\JSONnumbervaluestyle}%
    \fi
  \fi
  \lsthk@DetectKeywords%
}

\lst@AddToHook{EOL}%
  {\global\colonfoundonthislinefalse}

\makeatother

\usepackage{etoolbox}
\usepackage{natbib}
\usepackage{url}
\newcounter{bibcount}
\makeatletter
\patchcmd{\@lbibitem}{\item[}{\item[\hfil\stepcounter{bibcount}{[\thebibcount]}}{}{}
\renewcommand\NAT@bibsetup%
  [1]{\setlength{\leftmargin}{\bibhang}\setlength{\itemindent}{-\parindent}%
      \setlength{\itemsep}{\bibsep}\setlength{\parsep}{\z@}}
\makeatother

\begin{document}




\title{PC Agent: While You Sleep, AI Works - A Cognitive Journey into Digital World} 

\author{
    \textbf{Yanheng He}$^{1,3*}$\quad 
    \textbf{Jiahe Jin}$^{1,3*}$\quad \\
    \textbf{Shijie Xia}$^{1,2,3}$\quad 
    \textbf{Jiadi Su}$^{3}$\quad 
    \textbf{Runze Fan}$^{1,3}$\quad 
    \textbf{Haoyang Zou}$^{3}$\quad
    \textbf{Xiangkun Hu}$^{3}$\quad 
    \textbf{Pengfei Liu}$^{1,2,3\dagger}$ \\
    \\
  $^1$Shanghai Jiao Tong University \quad $^2$SII \\
    $^3$Generative AI Research Lab (GAIR)
}

\maketitle

\renewcommand{\thefootnote}{\fnsymbol{footnote}} 

\footnotetext[1]{Equal Contribution.}
\footnotetext[2]{Corresponding author.}

\thispagestyle{fancy}
\fancyhead{}
\lhead{\includegraphics[height=0.67cm]{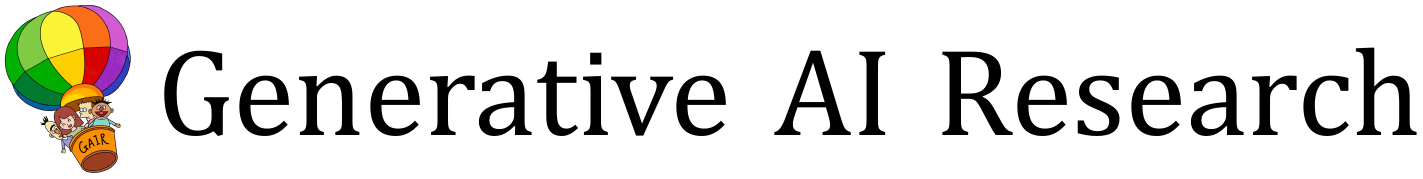}}
\renewcommand{\headrulewidth}{0pt}
\setlength{\headsep}{0mm}

\begin{abstract}

Imagine a world where AI can handle your work while you sleep - organizing your research materials, drafting a report, or creating a presentation you need for tomorrow. However, while current digital agents can perform simple tasks, they are far from capable of handling the complex real-world work that humans routinely perform. We present \textbf{PC Agent}, an AI system that demonstrates a crucial step toward this vision through \textbf{human cognition transfer}. Our key insight is that the path from executing simple ``tasks'' to handling complex ``work'' lies in efficiently capturing and learning from human cognitive processes during computer use. To validate this hypothesis, we introduce three key innovations: (1) PC Tracker, a lightweight infrastructure that efficiently collects high-quality human-computer interaction trajectories with complete cognitive context; (2) a two-stage cognition completion pipeline that transforms raw interaction data into rich cognitive trajectories by completing action semantics and thought processes; and (3) a multi-agent system combining a planning agent for decision-making with a grounding agent for robust visual grounding. 
Our preliminary experiments in PowerPoint presentation creation reveal that complex digital work capabilities can be achieved with a small amount of high-quality cognitive data - PC Agent, trained on just 133 cognitive trajectories, can handle sophisticated work scenarios involving up to 50 steps across multiple applications. This demonstrates the data efficiency of our approach, highlighting that the key to training capable digital agents lies in collecting human cognitive data.
By open-sourcing our complete framework, including the data collection infrastructure and cognition completion methods, we aim to lower the barriers for the research community to develop truly capable digital agents. Resources are available at \url{https://gair-nlp.github.io/PC-Agent/}.

\begin{figure}[H]
    \centering
    \includegraphics[width=0.9\linewidth]{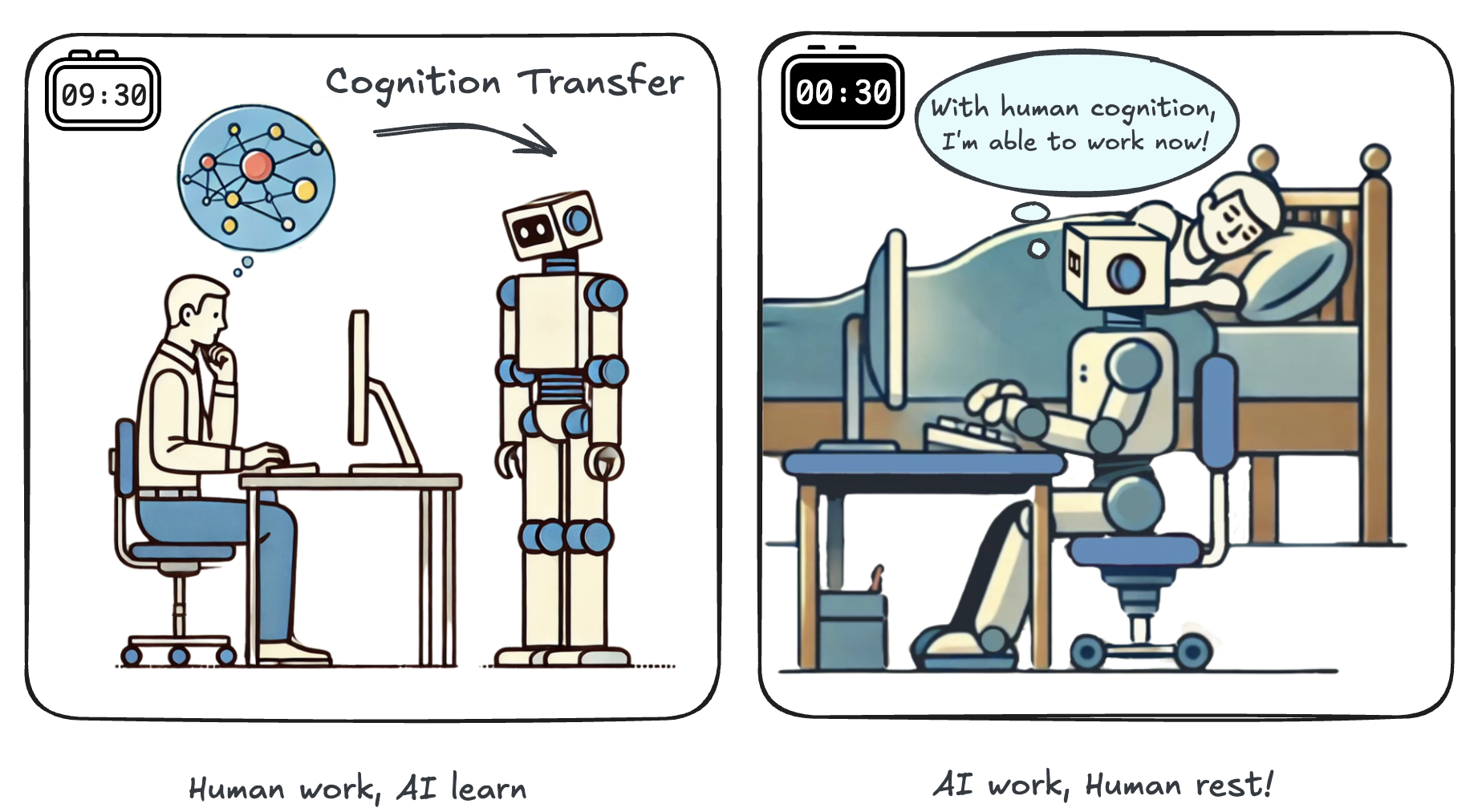}
\end{figure}

\begin{figure}
    \centering
    \includegraphics[width=1\linewidth]{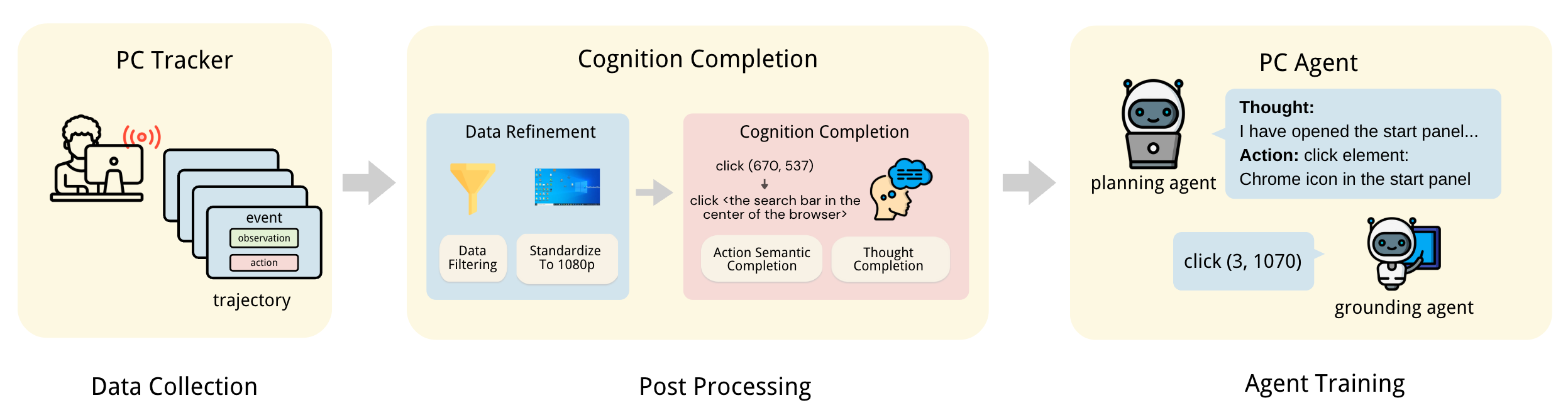}
    \caption{Overview of our framework, consisting of three key components: (1) PC Tracker, a lightweight infrastructure that collects human-computer interaction trajectories by recording user actions and state observations; (2) a two-stage cognition completion that converts raw interaction data into cognitive trajectories through data refinement and human cognition completion, including action semantics and thought processes; and (3) a multi-agent system comprising a planning agent for action decision-making and a grounding agent for click position grounding.}
    \label{fig:overview}
\end{figure}

\end{abstract}

\newpage

\pagestyle{fancy}
\lhead{\rightmark}
\renewcommand{\headrulewidth}{0.7pt}
\setlength{\headsep}{5mm}

\section{Introduction}


While artificial intelligence (AI) has made remarkable progress in understanding and generating content like text~\cite{chatgpt,openai2024gpt4technicalreport} and images~\cite{rombach2022highresolutionimagesynthesislatent, ramesh2022hierarchicaltextconditionalimagegeneration, li2022blipbootstrappinglanguageimagepretraining}, these capabilities exist largely in the representation world, where AI systems master pattern recognition and symbolic processing~\cite{bengio2014representationlearningreviewnew}. Recent advances in large language model (LLM) powered autonomous agents have garnered significant attention for their ability to perform task-oriented interactions~\cite{weng2023llmagents,wu2023autogenenablingnextgenllm}. In the domain of digital agents - AI systems that assist with digital tasks - current solutions have demonstrated basic capabilities in simple tasks like web searches~\cite{yao2023webshopscalablerealworldweb,deng2023mind2webgeneralistagentweb}. However, they remain far from capable of handling complex real-world work that humans routinely perform~\cite{xie2024osworldbenchmarkingmultimodalagents, bonatti2024windowsagentarenaevaluating}, such as video editing, presentation creation, or report writing - activities that require sustained operation across multiple applications, sophisticated decision-making, and even aesthetic judgment. To meaningfully reduce human workload, digital agents must evolve beyond simple task execution to manage more complex work efficiently.

\paragraph{The Critical Gap: From Simple Task Automation to Complex Work Completion}
Most existing approaches in building digital agents rely heavily on proprietary LLM APIs~\cite{wu2024oscopilotgeneralistcomputeragents,zheng2024gpt4visiongeneralistwebagent,zhang2023appagentmultimodalagentssmartphone}. While few works have attempted to train models for computer operation, these efforts have primarily concentrated on either improving basic visual grounding capabilities \cite{gou2024navigatingdigitalworldhumans, wu2024osatlasfoundationactionmodel} or specific domains like web \cite{liu2024harnessingwebpageuistextrich}. However, these approaches still struggle with complex real-world computer work. Through our analysis, we identify two critical challenges:
(1) \textbf{foundational visual grounding}:
The ability to precisely locate GUI elements (e.g. the Start button in the taskbar) represents a fundamental capability that current vision-language models still struggle with~\cite{xie2024osworldbenchmarkingmultimodalagents, gou2024navigatingdigitalworldhumans}.
(2) \textbf{complex cognitive understanding}:
More crucially, current agents lack the cognitive capabilities required for complex work. They struggle with maintaining contextual awareness across extended interaction sequences, making dynamic decisions based on changing environments, and adapting strategies in response to execution outcomes. While prompt engineering~\cite{liu2021pretrainpromptpredictsystematic} can partially help, it proves inadequate for truly complex work.

\paragraph{Human Cognition: The Missing Key}

Our key insight is that the path to the digital world lies in \textbf{human cognition transfer}. When completing complex work, the human brain engages in sophisticated cognitive activities - understanding objectives, analyzing current states, reflecting on past actions, and planning future strategies. These cognitive processes ultimately crystallize into clear decisions, which are then externalized into observable behavior.
This observation leads us to develop a novel framework that efficiently captures and transfers human cognition to AI agents.

\paragraph{Our Contributions}

In this work, we present three key contributions that together form a complete solution for developing truly capable digital agents:

\begin{itemize}
    \item \textbf{PC Tracker}: We introduce the first open-source infrastructure for efficiently collecting large-scale human-computer interaction trajectories. This lightweight system captures not just actions but the complete cognitive context of human computer use, establishing a rich foundation for agent training.
    \item \textbf{Cognition Completion Pipeline}: We develop an innovative approach that transforms raw interaction data into cognitive trajectories by completing action semantics and thought processes in steps. This enables AI to learn not just what humans do, but why and how they make decisions in complex digital environments.
    \item \textbf{PC Agent}: We propose an AI system that can perform complex cognitive work in the digital world. Using PowerPoint presentation creation as our main testing ground, PC Agent shows remarkable capabilities in executing long-sequence tasks and switching between applications, allowing it to create complex presentations with practical utility.
\end{itemize}

\paragraph{Findings and Impact}

Our preliminary experiments demonstrate \textbf{significant data efficiency} in learning from human cognitive processes rather than just behavioral data - PC Agent, trained on just 133 cognitive trajectories, can execute complex work with up to 50 steps. By \textbf{open-sourcing our framework}, we provide the research community with more than just a proof of concept; we offer a practical foundation for further research and development. As we progress toward a broader vision - from representation world to digital world, and ultimately to the physical world - this work represents a critical step in enabling AI to practically enhance human productivity.

\section{How Far Are We From True Digital Agents}


\subsection{The Capability Gap in Real-world Applications: From ``task'' to ``work''}

In the evolution of digital agents, LLM-powered autonomous agents have garnered significant attention due to their ability to engage in complex task-oriented interactions~\cite{weng2023llmagents,wu2023autogenenablingnextgenllm}. Early works like ReAct \cite{yao2023reactsynergizingreasoningacting} and Reflexion \cite{shinn2023reflexionlanguageagentsverbal} established fundamental frameworks for agent reasoning and self-optimization. In implementing digital agents, two primary technical approaches emerged: backend access \cite{trivedi2024appworldcontrollableworldapps, openinterpreter} and frontend GUI interaction. The latter has gained prominence by operating without backend access permissions, offering universal applicability, better security, and adaptability to interface changes. With the advancement of vision-language models \cite{liu2023visualinstructiontuning,gpt-4vreport,openai2024gpt4o,wang2024qwen2vlenhancingvisionlanguagemodels}, GUI agents have made significant progress in both visual understanding \cite{hong2023cogagentvisuallanguagemodel,
liu2024harnessingwebpageuistextrich,
you2024ferretuigroundedmobileui,
yang2023setofmarkpromptingunleashesextraordinary, cheng2024seeclickharnessingguigrounding,
lu2024omniparserpurevisionbased} and general task capabilities, demonstrated by applications across web browsers \cite{gur2024realworldwebagentplanninglong,
zheng2024gpt4visiongeneralistwebagent,
yao2023webshopscalablerealworldweb,nakano2022webgptbrowserassistedquestionansweringhuman}, mobile devices \cite{zhang2023appagentmultimodalagentssmartphone, hoscilowicz2024clickagentenhancinguilocation, zhang2024androidzoochainofactionthoughtgui, zhang2024lookscreensmultimodalchainofaction} and desktops \cite{wu2024oscopilotgeneralistcomputeragents, zhang2024ufouifocusedagentwindows}. To advance the field, researchers have established important evaluation benchmarks through comprehensive datasets \cite{rawles2023androidwildlargescaledataset, deng2023mind2webgeneralistagentweb} and execution-based environments that simulate real-world scenarios \cite{zhou2024webarenarealisticwebenvironment, koh2024visualwebarena, xie2024osworldbenchmarkingmultimodalagents, bonatti2024windowsagentarenaevaluating, rawles2024androidworlddynamicbenchmarkingenvironment}.

Despite notable progress in the field, we remain far from achieving truly capable digital agents. Even the most advanced systems, including the recently celebrated new Claude-3.5-Sonnet \cite{anthropic2024models}, still significantly under-perform humans in computer use \cite{xie2024osworldbenchmarkingmultimodalagents,hu2024dawnguiagentpreliminary}. While current digital agents can perform simple \textbf{tasks} like web searches and file copying, they face significant challenges when tackling comprehensive \textbf{work} that better reflect real-world computer use. Consider video editing, presentation creation, and report generation - these work require sustained operation across multiple applications, sophisticated decision-making, and even human-level aesthetic judgment. We argue that true digital agents should be able to efficiently handle these complex work, not just simple tasks, to meaningfully reduce human workload.

\subsection{Breaking Barriers: Visual Grounding and Cognitive Understanding}
\label{sec: barriers}

To realize true digital agents, we have identified two critical technical challenges: foundational visual grounding capabilities and deep cognitive understanding.

\paragraph{Visual Grounding}
\label{sec:visual_grounding}

Visual grounding - the ability to precisely locate elements in GUI - represents a fundamental capability for agents to effectively use computers. This is because click-based interactions, which require precise coordinate outputs, constitute a significant portion of computer tasks. However, most current vision-language models (VLMs), including state-of-the-art proprietary models like GPT-4o and open-source models like Qwen2-VL, still lack this basic capability. Contemporary works such as UGround \cite{gou2024navigatingdigitalworldhumans} and OS-ATLAS \cite{wu2024osatlasfoundationactionmodel} attempt to address this issue by fine-tuning models on large-scale GUI visual grounding datasets. While successful in developing the target capability, these approaches compromise the model’s general question-answering and instruction-following abilities. Notably, Anthropic's new Claude-3.5-Sonnet has emerged as the first frontier model with state-of-the-art computer use capabilities, though its training details remain undisclosed. We believe that both specialization and general capabilities are equally important for effective agent models. Encouragingly, we discover that the open-source general-purpose VLM Molmo \cite{deitke2024molmopixmoopenweights} demonstrates exceptional visual grounding abilities, establishing a strong foundation for open-source digital agents.


\paragraph{Cognitive Understanding}
\label{cognition understanding}

However, even with robust visual grounding capabilities, current agents still struggle to complete complex computer work. Our analysis reveals that cognitive understanding represents the crucial missing piece in this puzzle - one that we will address through our human cognition transfer framework. This limitation manifests in two key aspects:

\begin{itemize}
    \item \textbf{{Lack of Fine-grained Cognitive Knowledge}}: 
While current agents can perform high-level task planning, they lack the fine-grained cognitive knowledge for specific GUI operations. For instance, adding a title in PowerPoint requires clicking the title box before typing. Such operational sequences, intuitive to humans, present challenges for agents. 
Moreover, when handling cross-application tasks, such as collecting images from browsers to presentations, agents must comprehend the interaction logic and division of labor between different applications. This challenge necessitates reconsidering agent training from a human cognitive perspective, as GUIs were fundamentally designed for human interaction.
    \item \textbf{{Insufficient Agent Training}}:
While current language models exhibit strong capabilities in following instructions and generating responses, they often struggle with the unique demands of agent workflows. These workflows require maintaining contextual awareness across extended interaction sequences, making dynamic decisions based on changing environments, and adapting strategies in response to execution outcomes. Although prompt engineering can partially bridge this gap, it proves inadequate for truly complex work. This misalignment between model capabilities and agent requirements highlights the need for specialized training approaches to enable effective real-world work completion.
\end{itemize}

\section{Cognition Transfer: A Gateway for AI into the Digital World}

The journey to the digital world represents a fundamental evolution from understanding and responding in language to taking meaningful action for LLMs. Although current digital agents can perform simple tasks, they are still far from effectively completing real-world work in digital environments, where humans manage their everyday work and life activities. The fundamental challenge lies here: \textit{how can we enable AI systems to fundamentally evolve from understanding tasks to executing them in the digital world?}

We posit that \textbf{human cognition transfer} is pivotal in addressing this challenge. Human cognition is primarily based on the dynamic interaction between internal thinking and external behavior. When performing complex work in real environments, the human brain engages in sophisticated cognitive activities, including understanding objectives, analyzing current states, reflecting on the past, and planning future strategies. This series of cognitive processes ultimately crystallizes into clear decisions, which are then externalized into observable behavior. In this sense, behavior serves as the external projection of complex cognitive activities in the human brain.

If AI systems can acquire human cognition to interact with the digital world, they can complete complex work as naturally as humans do. However, current technology cannot directly record cognitive activities in the brain. To address this, we propose an indirect approach:

\begin{enumerate}
    \item First, we designed a lightweight infrastructure to \textbf{efficiently collect raw interaction trajectories} between humans and digital devices. These trajectories not only capture the specific steps of task execution but also reflect dynamic exploration, trial and error, and optimization. 
    \item Then, we leverage LLMs to analyze these raw trajectories and \textbf{complete the cognitive processes} behind human behavior by completing action semantics and thought processes in steps, transforming them into \textit{cognitive trajectories}. These cognitive trajectories can be considered effective approximations of genuine human cognitive activities.
\end{enumerate}

By learning from these cognitive trajectories, AI systems can not only mimic specific operational behaviors but also master the underlying human cognition more efficiently. Our experimental results demonstrate remarkable data efficiency in enabling AI systems to perform complex computer work through human cognition transfer.

\section{PC Tracker: Human-Computer Interaction Data Collection Infrastructure}

\begin{wrapfigure}{r}{0.32\textwidth}
    \centering
    \includegraphics[width=0.92\linewidth]{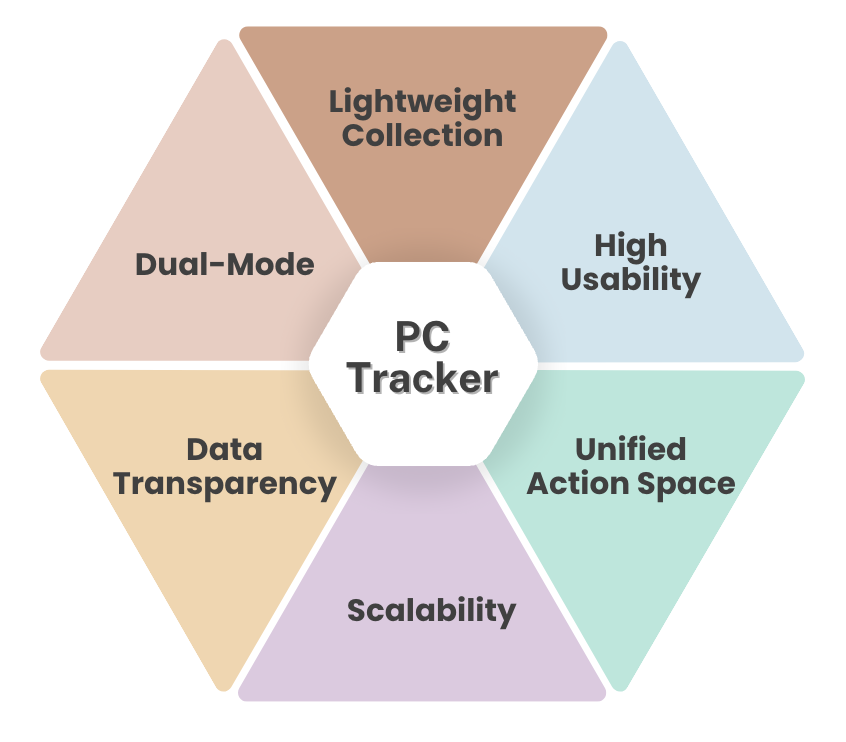}
    \caption{Key features of PC Tracker}
    \label{fig: key feature}
\end{wrapfigure}

While supervised fine-tuning has proven effective for adapting LLMs to many tasks, large-scale computer interaction trajectories remain severely limited compared to conventional text or image datasets. This data bottleneck has become a key impediment in developing digital agents that can effectively use computers like humans do.

\begin{figure}
    \centering
    \includegraphics[width=1\linewidth]{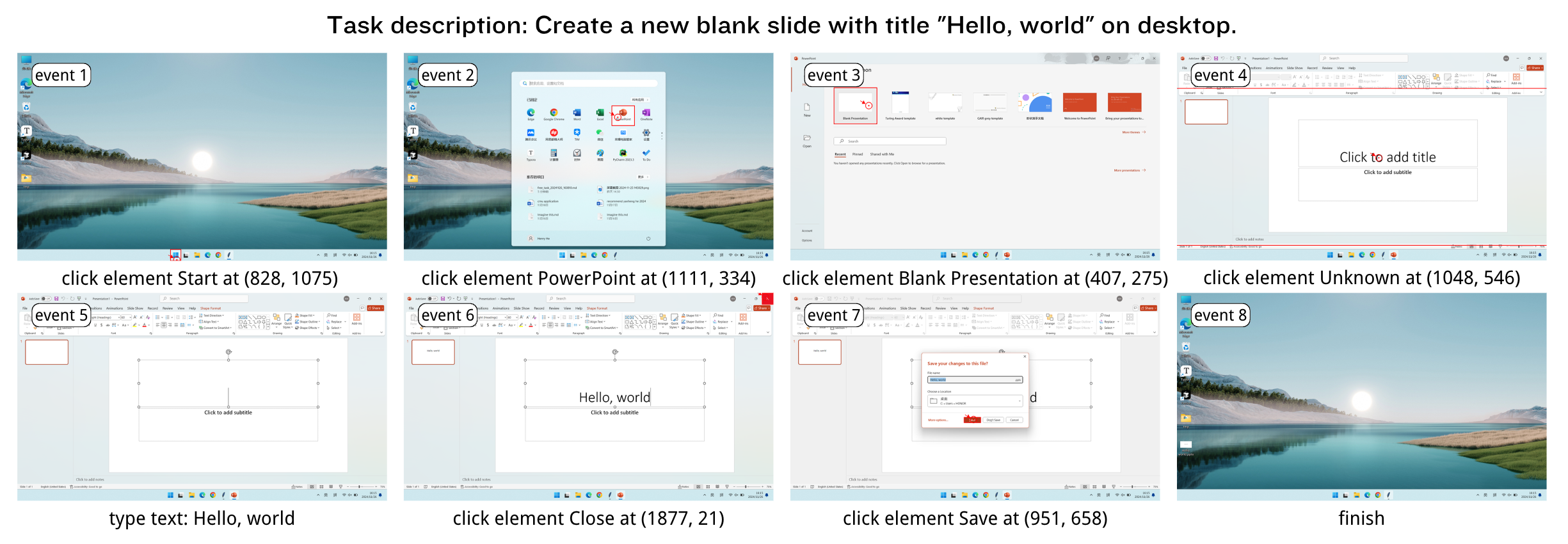}
    \caption{An example trajectory collected by PC Tracker. Red marks on the screenshots indicate the positions of click-related actions.}
    \label{fig: trajectory example}
\end{figure}

To address this bottleneck, we present PC Tracker, the first lightweight infrastructure for efficient large-scale collection of human-computer interaction data. Similar to screen recording mechanisms, PC Tracker runs seamlessly in the background, recording user actions by monitoring keyboard and mouse activities while capturing screenshots to document state observations. This enables the collection of real-world interaction trajectories (see an example in Figure~\ref{fig: trajectory example}) at scale, establishing a rich foundation for future research in agent training and cognition engineering. PC Tracker is designed with the following key features, as summarized in Figure~\ref{fig: key feature}.

\paragraph{Lightweight Collection}

\setlength{\columnsep}{15pt}  
\setlength{\intextsep}{0pt}   

\begin{wrapfigure}{r}{0.4\textwidth}
\centering
\small
\resizebox{\linewidth}{!}{%
\begin{tabular}{@{}lp{5.5cm}@{}}
\toprule
\textbf{Action} & \textbf{Description} \\ \midrule
\textit{click (x, y)} & clicks at coordinates. \\
\textit{right click (x, y)} & right-click at coordinates. \\
\textit{double click (x, y)} & double-click at coordinates. \\
\textit{press (x, y)} & press mouse down at coordinates. \\
\textit{drag to (x, y)} & drags the mouse to coordinates. \\
\textit{scroll (0, 10)} & scrolls the screen with offset dy = 10. \\
\textit{press key: enter} & presses the Enter key. \\
\textit{hotkey (ctrl, c)} & performs the Ctrl+C hotkey (copy). \\
\textit{type text: hello} & type text ``hello''. \\
\textit{wait} & pauses for some time. \\
\textit{finish} & the task is finished. \\
\textit{fail} & the task is failed. \\ \bottomrule
\end{tabular}%
}
\caption{Action space $\mathcal{A}$ of PC Tracker.\newline}
\label{tab:action-space}
\end{wrapfigure}

Unlike bulky and redundant video recording solutions, PC Tracker efficiently collects data through event-based tracking. It captures critical events \( e = \langle act, obs\rangle \) when user operation is detected, where \( act \) represents action and \( obs \) represents the corresponding computer state observation. Critical events compose the trajectory, capturing the complete and authentic human-computer interaction while significantly reducing storage usage.

\paragraph{High Usability}
Running seamlessly in the background, PC Tracker allows users to use their computers naturally. Unlike some contemporary approaches that record complete accessibility trees, we deliberately avoid this practice as the crawling process would introduce noticeable latency and disrupt user operations.

\paragraph{Scalability} 
The lightweight design and high usability ensure the feasibility of large-scale, long-term deployment, enabling PC Tracker to support unlimited-scale data collection. Our statistics show that an hour of continuous computer usage typically generates around 2,000 events, demonstrating immense data potential.

\paragraph{Unified Action Space} 
We encapsulate fragmented keyboard and mouse operations into a unified action space. Raw operations are combined into actions like double click and type, significantly reducing the number of recorded events while enhancing action semantics. 

\paragraph{Data Transparency} 
We ensure user privacy through local data storage, complete recording control, and transparent Markdown visualization of all stored trajectories.

\paragraph{Dual-Mode}
PC Tracker offers two recording modes: task oriented and non-task oriented. The non-task oriented mode captures user interaction trajectories without specific tasks, ideal for large-scale data collection, while the task oriented mode primarily serves for supervised fine-tuning data annotation.

\subsection{Tracking Strategy}

\subsubsection{Action Recording}
\label{sec:action-recording}

The original records of certain actions are often fragmented, leading to a loss of semantic information, such as double-click, scroll, hotkey, and type. For instance, a type action is split into individual key press operations. Besides, some actions require special rules to be identified from the original operations, such as the \textit{drag to} action, which is originally just a normal mouse release.

\begin{wrapfigure}{r}{0.4\textwidth}
\includegraphics[width=0.95\linewidth]{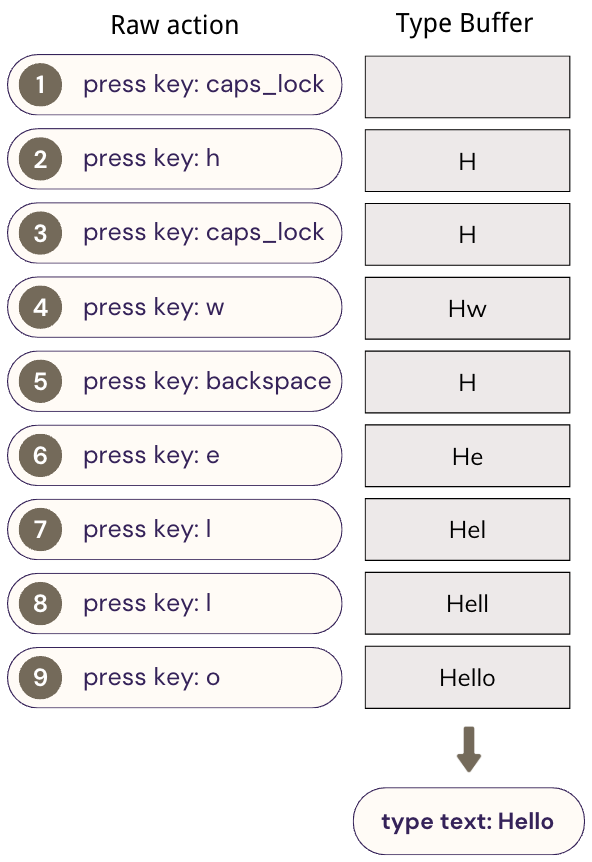}
\caption{Example of type encapsulation.\newline}
\label{fig: type encapsulation}
\includegraphics[width=1\linewidth]{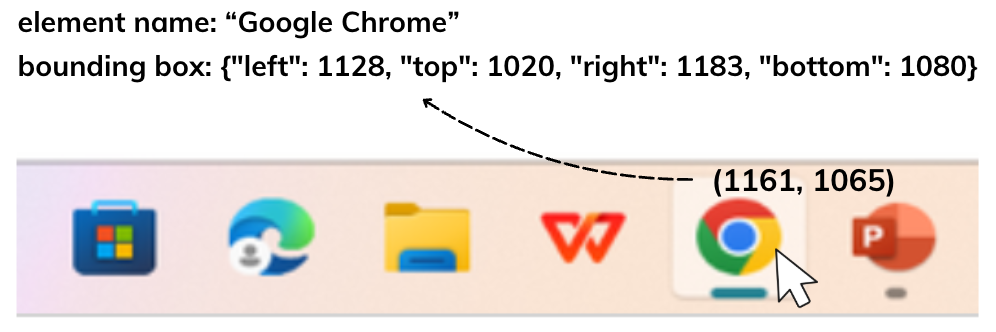}
\caption{An example of the output from $\operatorname{get\_element\_info\_at\_position}(x, y)$ when the user clicks Chrome icon at $(1161, 1065)$.\newline}
\label{fig:get_element}
\includegraphics[width=1\linewidth]{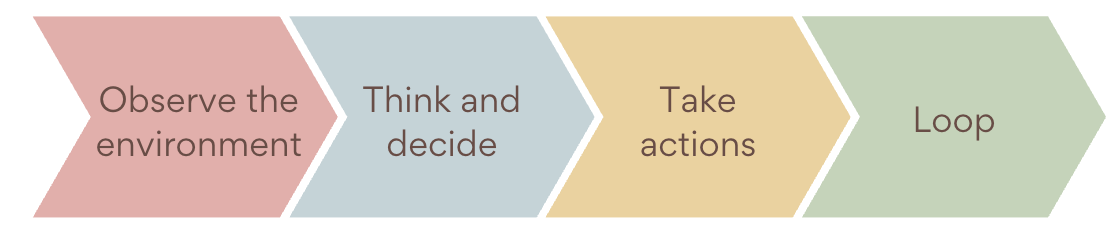}
\caption{Natural flow of human interaction: Observe, Think, Act.}
\label{fig: natural flow}
\end{wrapfigure}

To address this issue, we designed heuristic algorithms that track the history of mouse and keyboard activities to encapsulate these raw operations into a unified action space $\mathcal{A}$ designed based on OSWorld \cite{xie2024osworldbenchmarkingmultimodalagents}, as shown in Figure~\ref{tab:action-space}. This action space, originally designed for AI agents, significantly reduces the difficulty for AI to understand and learn human operational behaviors. For example, Figure~\ref{fig: type encapsulation} demonstrates how PC Tracker encapsulates 9 raw actions into a single \textit{type text} action when the user intends to type ``Hello”. The user’s raw actions involve capitalization change, character typing, and error corrections, as shown in the left column. These raw actions are
accumulated into the type buffer and combined into the unified action ``type text: Hello” upon detecting type completion.

\setlength{\columnsep}{15pt}  
\setlength{\intextsep}{10pt}   

Additionally, we observed that coordinate-based click-related actions (the top five actions in Figure~\ref{tab:action-space}) lack sufficient semantic information compared to other actions like keyboard inputs. For instance, \textit{click (333, 444)} is far more abstract than \textit{press key: enter}. To support downstream action semantic completion, PC Tracker records additional contextual information for click-related actions. Specifically, we implemented a low-overhead function $\operatorname{get\_element\_info\_at\_position}(x, y)$ that efficiently retrieves the element's information from coordinates via system API, including its bounding box coordinates and element name (see Figure~\ref{fig:get_element}). When a click-related action occurs, PC Tracker passes coordinates to this function and records the clicked element's information.

\subsubsection{Observation Capture}

We adopt a carefully designed timing strategy to record corresponding screenshots as state observations for actions. While it may seem intuitive to record the screenshot at the moment an action is detected, this can lead to inaccurate observation records. Figure~\ref{fig: natural flow} shows that humans naturally observe the environment state before making decisions and taking actions. Therefore, PC Tracker records the screenshot just before the action occurs. To achieve this, our system continuously takes screenshots and keeps the latest one in memory. When an action is detected, this latest cached screenshot (approximately 0.1 seconds prior) is recorded as the observation.

The design of only recording critical events offers notable advantages over complete video recording. It significantly reduces storage requirements, thus allowing us to capture screenshots at maximum resolution - a crucial feature for future models, as many interface details such as small-font text are only legible at high resolutions.

Besides, while many works require a complete accessibility tree to aid observation \cite{jia2024agentstorescalableintegrationheterogeneous, agashe2024agentsopenagentic}, we find this impractical in real use. Crawling the accessibility tree requires a lengthy recursive process (taking tens of seconds to minutes), which forces users to wait after each action until the tree traversal completes, severely disrupting normal computer usage. Given the rapid progress in VLMs, we believe maintaining such exhaustive tree records is unnecessary.

\subsection{Dual-Mode Collection}

\begin{wrapfigure}{r}{0.55\textwidth}
    \centering
    \includegraphics[width=1\linewidth]{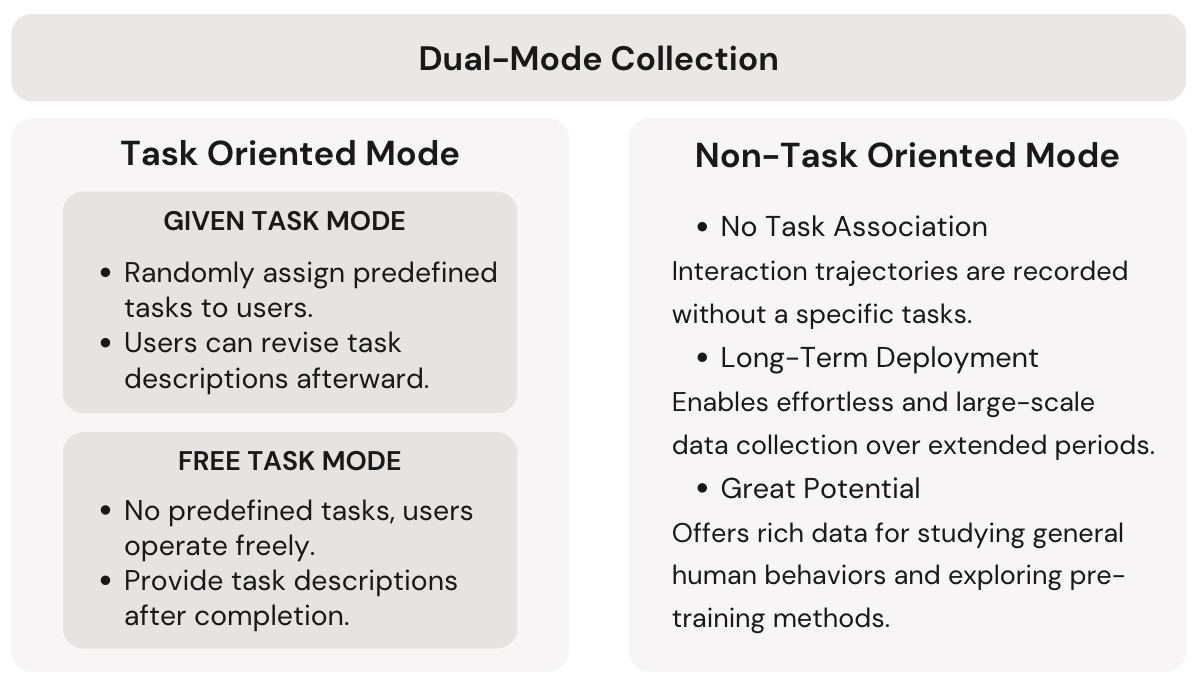}
    \caption{An overview of the dual-mode collection design}
    \label{fig:dual-mode}
\end{wrapfigure}

PC Tracker supports two recording modes: task oriented and non-task oriented, primarily differentiated by whether the interaction trajectories are associated with certain task descriptions. An overview is shown in Figure~\ref{fig:dual-mode}.

The task oriented mode records interaction trajectories that are explicitly associated with specific tasks, primarily designed to support easy annotation for supervised fine-tuning. It is divided into given task and free task modes. In the given task mode, PC Tracker randomly assigns tasks from a predefined task library to users, who then use the
computer to complete the tasks while their interaction trajectories are recorded. In the free task mode, users can freely use the computer without predefined constraints and provide a task description for their operation sequence after completion. Notably, even in the given task mode, users have the flexibility to revise task descriptions after stopping the recording. This feature allows users to more accurately describe their activities, particularly if their operations deviated from the original assigned task or if they discovered a better way to characterize their work.

In the non-task oriented mode, the system directly records interaction trajectories without a specific task, enabling long-term deployment and effortless collection of large-scale data. While this mode may require additional approaches to infer user intentions, it provides a rich foundation for studying general human operational behaviors and exploring pre-train methods.

\subsection{Privacy and Data Transparency}
PC Tracker integrates privacy considerations and data transparency in its design. During recording sessions, users have full control over their data - if they notice sensitive information has been captured or if they are not satisfied with their task completion process, they can choose to discard the recording immediately without saving it. For saved trajectories, PC Tracker stores recordings locally and visualizes them in Markdown format, allowing users to easily review and check their recorded operations at any time.

\section{Cognition Completion: From Raw Interaction Data to Cognitive Trajectory}

Converting raw human-computer interaction data into meaningful cognitive trajectories is essential for AI to learn from human demonstrations. To achieve this goal, we conduct two-stage sequential post-processing: data refinement to ensure quality, followed by cognition completion to extract atomic action semantics and reconstruct the underlying thought process for each action.

\subsection{Data Refinement}
To optimize the quality of raw interaction data, we designed and implemented a comprehensive pipeline that consists of three steps: 1) trajectory filtering, 2) action filtering, and 3) standardization.

\paragraph{Trajectory filtering}
First, we examined the completeness of trajectories and files to eliminate error data caused by unexpected incidents (such as forced termination of PC Tracker), thereby ensuring robustness.

\paragraph{Action filtering}
Next, we focused on identifying and removing two categories of actions: (1) actions associated with PC Tracker, such as clicks on the start record button; (2) redundant actions in actual human operations, like control key presses for hotkey prefixes, consecutive waits due to operation pauses, and meaningless clicks. We found that these actions not only introduce unnecessary burdens to model learning but also interfere with the subsequent cognition completion process.

\paragraph{Standardization}
Finally, to ensure consistency across different data sources, we standardized all screenshot resolutions to 1080p (1920×1080), providing a standardized image input format for training.

\subsection{Cognition Completion}

\begin{figure}[h]
    \centering
    \includegraphics[width=1\linewidth]{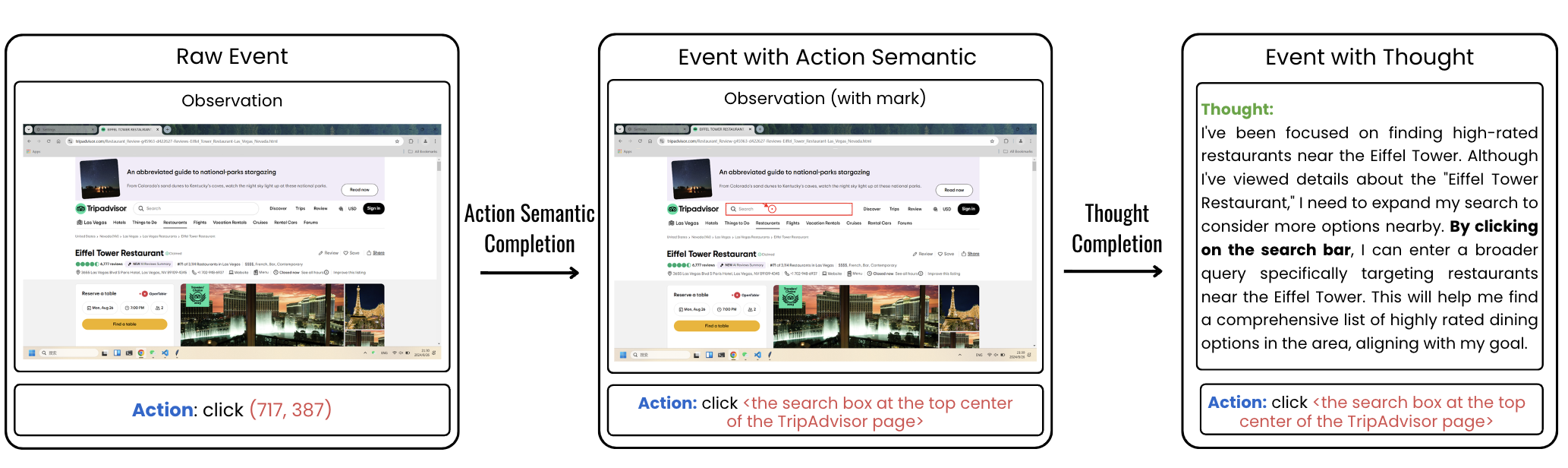}
    \caption{Visualization of our cognition completion process for a click action. (Left) Raw click event recorded by PC Tracker. (Center) Action semantic completion converts coordinates (717, 387) into a semantic description ``the search box at the top center of the TripAdvisor page". (Right) Thought completion reconstructs the human intention behind this action - finding high-rated restaurants near the Eiffel Tower by broadening the search scope.}
    \label{fig:cognition-completion}
\end{figure}

Building upon the refined trajectories, the cognition completion stage tackles a more crucial challenge: reconstructing human cognition of computer use. This stage consists of two steps: the first step focuses on reconstructing the atomic semantics of actions, and the second step reconstructs the thought processes behind actions based on the enriched action semantics. A detailed view is shown in Figure~\ref{fig:cognition-completion}.

\subsubsection{Action Semantic Completion}
\label{sec:action-completion}

As mentioned in Section \ref{sec:action-recording}, click-related actions lack semantic information due to their coordinate-based nature. To comprehensively reconstruct human cognition, we supplement click-related actions with semantic information before thought completion by generating high-quality descriptions for click targets.

PC Tracker records additional information for click elements, including the bounding box coordinates and element name (as shown in Figure~\ref{fig:get_element}). We highlight the click position and bounding box with red marks in the corresponding screenshot, which visualizes click-related actions on the screenshot to facilitate model understanding. This marked screenshot serves as visual input in both action semantic completion and thought completion phases. 

When generating descriptions for click targets, we leverage GPT-4o to take the marked screenshot and the element name as input and generate high-quality descriptions. Notably, while PC Tracker records element names through system API, these names are often inadequate: many are missing, overly generic (e.g., numerous elements in Chrome simply labeled as ``Chrome''), excessively verbose (e.g., containing complete text content for text elements), or even incorrect. In contrast, GPT-4o generates accurate and concise descriptions of click targets, facilitating the understanding of actions during thought completion.

\subsubsection{Thought Completion}

After completing click element information, the thought completion step reconstructs the implicit reasoning behind human actions using an iterative generation approach based on previously obtained information. Specifically, for each action in the trajectory, we provide GPT-4o with multi-dimensional input information: the task description, preceding actions with their associated thoughts, subsequent actions after semantic completion, and the corresponding marked screenshot mentioned in Section~\ref{sec:action-completion}. 

We found this comprehensive temporal context spanning past, present, and future, combined with enhanced visual information, is crucial for VLMs to understand and reconstruct human decision-making processes. Based on all this information, GPT-4o can effectively complete the accurate thought process underlying human actions, including task progression awareness, environmental exploration behaviors, and error recovery strategies.

\section{PC Agent: Cognitive Agent for Complex Computer Work}

This section presents our implementation of PC Agent, an AI system capable of complex computer work. Notably, our system is \textbf{built entirely on open-source models}, avoiding dependence on frontier proprietary models like GPT-4o or Claude-3.5-Sonnet.

\subsection{Multi-Agent System}

\begin{figure}[h]
    \centering
    \includegraphics[width=0.8\linewidth]{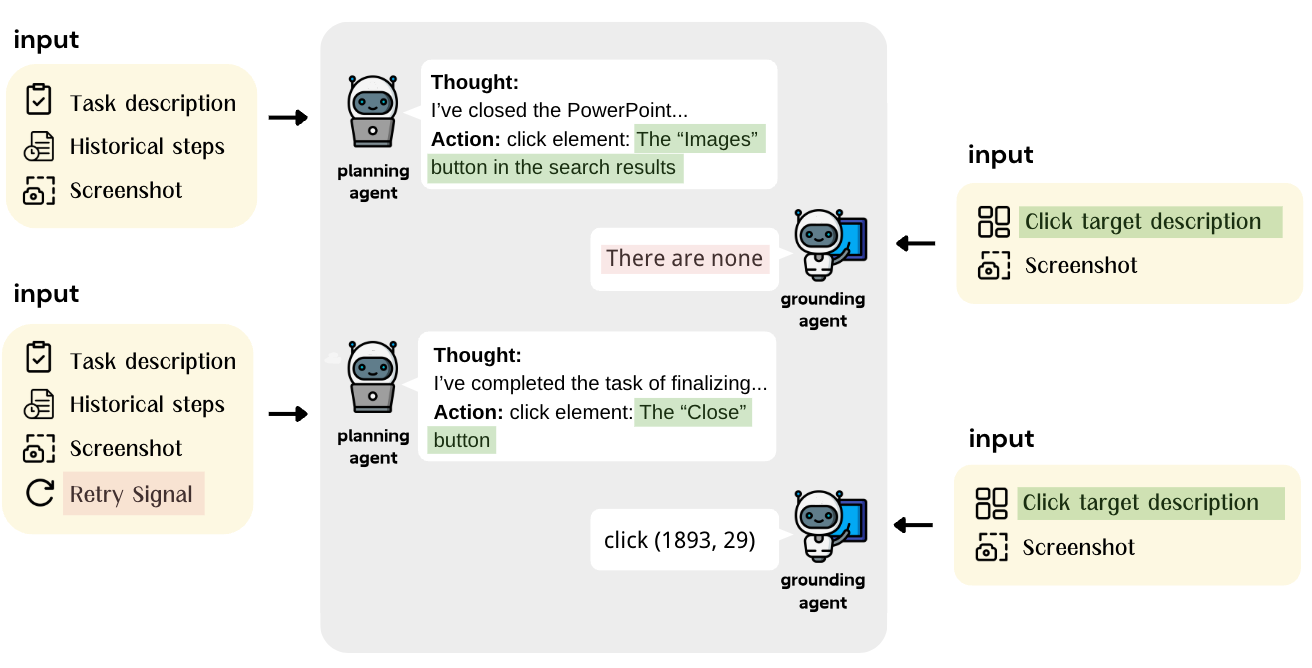}
    \caption{Illustration of multi-agent workflow. The planning agent initially attempts to click a nonexistent element \textit{The `Images' button}, which is reported by the grounding agent. Upon receiving this feedback, the planning agent reformulates its plan, and the grounding agent generates coordinates of the new click target. The workflow illustrates the error correction mechanism between agents.}
    \label{fig: collab}
\end{figure}

In Section~\ref{sec: barriers}, we identified two major challenges of current digital agents: visual grounding and cognitive understanding. In our preliminary implementation of PC Agent, we address these challenges through a multi-agent architecture. First, we trained a planning agent by learning from human cognitive trajectories, enabling it to acquire human cognition for effective planning. Second, we implemented a robust grounding agent with a self-validation mechanism that achieves
near-human perfection. The two agents work collaboratively: the planning agent handles action decision-making, while the grounding agent executes click-related actions.

The workflow proceeds as follows, illustrated with an example in Figure~\ref{fig: collab}: The planning agent first analyzes the task and observes the current state to generate a thought process and an action decision. If the action is not click-related, it will be directly executed. Otherwise, the generated target description of the click-related action will be forwarded to the grounding agent. The grounding agent not only generates the specific position coordinates of the click target but also validates grounding accuracy. If the grounding agent discovers that the planning agent is attempting to click on a non-existent target on the screen, the planning agent is prompted to reformulate its action plan. All actions are executed using the \texttt{PyAutoGUI} library.

We believe that at the current stage, the multi-agent architecture can better leverage existing VLMs' capabilities through clear task division and is likely to achieve better results in the short term. However, looking ahead, end-to-end systems like the new Claude-3.5-Sonnet may represent the future direction of this field.

\subsection{Training Planning Agent via Cognitive Trajectories}

\begin{wrapfigure}{r}{0.4\textwidth}
    \centering
    \includegraphics[width=1\linewidth]{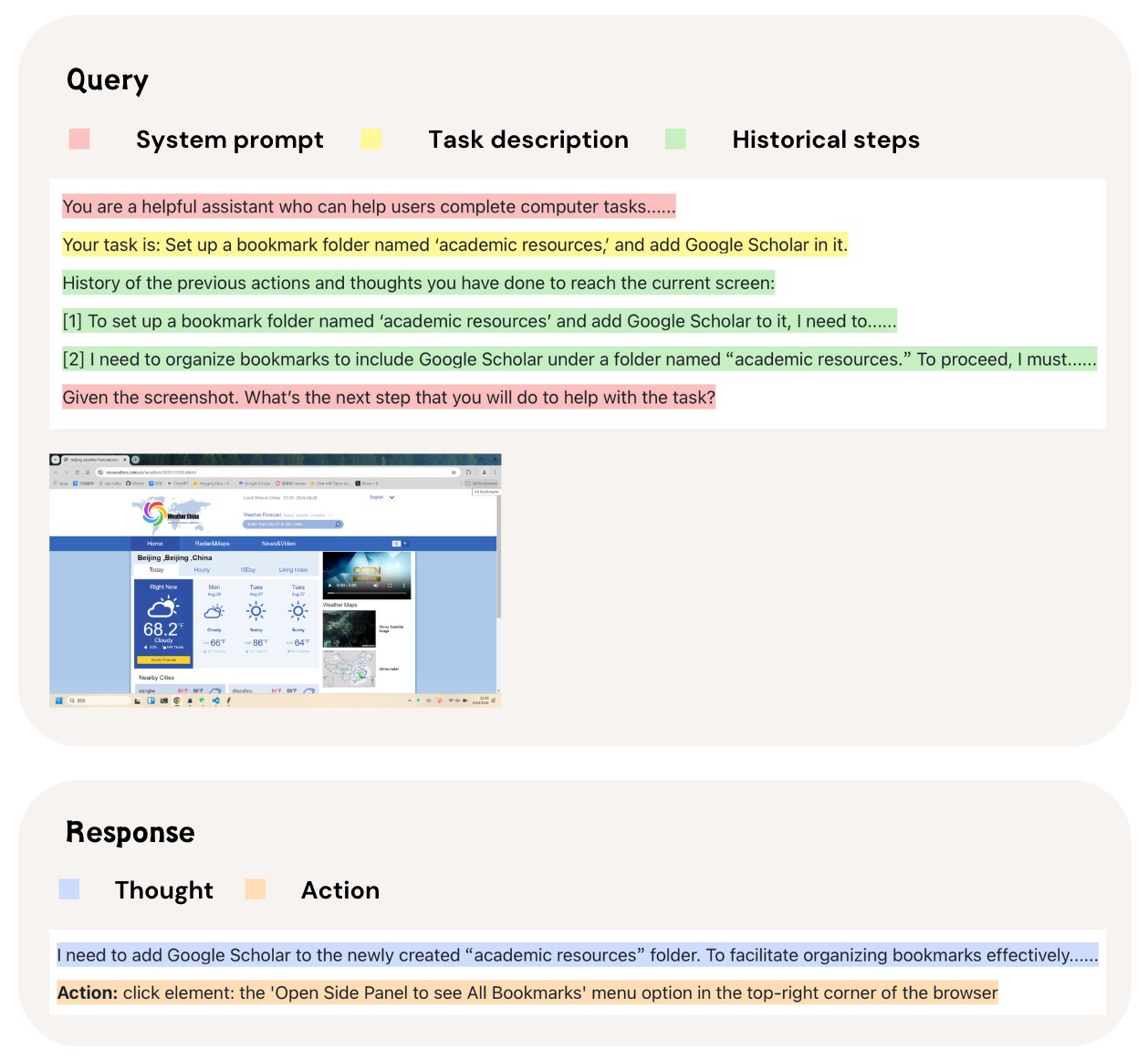}
    \caption{Training data example showing query and response structure.}
    \label{fig:train-data}
\end{wrapfigure}

Section~\ref{cognition understanding} posits that the limitations of current digital agents primarily stem from the lack of essential agent training data. To tackle this challenge, we utilize PC Tracker and cognition completion post-processing to efficiently build high-quality cognitive trajectories, which are then used to train the planning agent.

In the resulting training data (as shown in Figure~\ref{fig:train-data}), system prompts, task descriptions, and historical steps are used as textual inputs, while the corresponding screenshot (without mark) is used as the visual input. Note that each historical step consists of the thought process and action decision, incorporating the progress state within the overall task and any exploration behaviors from prior steps in natural language. For example, it allows the model to determine whether the desktop indicates the task has just started, or has been completed with all applications closed.

Furthermore, we structure the responses into distinct thought and action components. As discussed earlier, the thought process captures human cognition of computer use. This data organization enables the model to learn the underlying cognitive patterns behind actions and aligns with the ReAct \cite{yao2023reactsynergizingreasoningacting} paradigm where models generate verbal reasoning traces and actions in an interleaved manner.

\subsection{Robust Visual Grounding with Self-Validation}

\begin{figure}[h]
    \centering
    \includegraphics[width=0.85\linewidth]{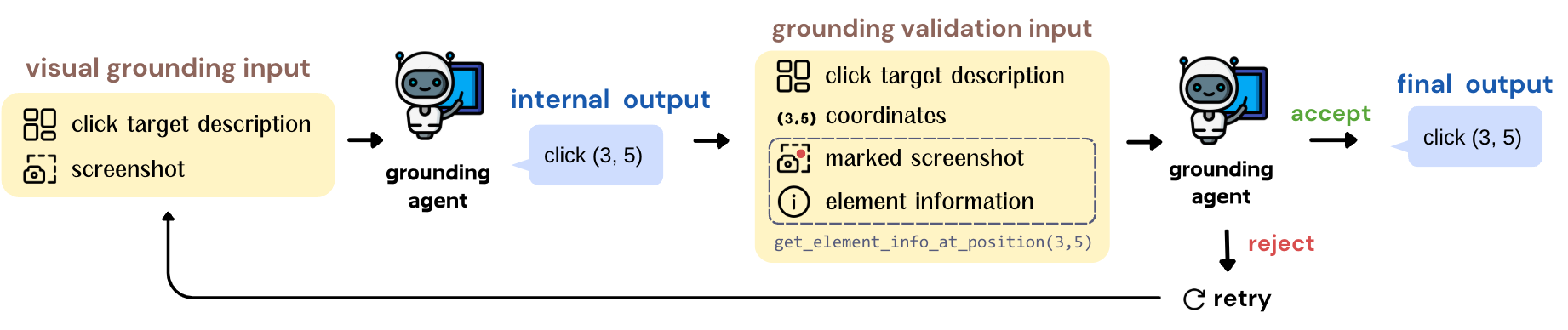}
    \caption{Illustration of the grounding agent’s self-validation mechanism.}
    \label{fig: grounding check}
\end{figure}

As outlined in Section~\ref{sec:visual_grounding}, visual grounding represents a fundamental capability for digital agents. While our exploration reveals that Molmo \cite{deitke2024molmopixmoopenweights} demonstrates remarkable proficiency in visual grounding, it still occasionally exhibits errors, particularly in the GUI domain where elements are densely arranged. However, even minor misalignments in visual grounding can lead to catastrophic failures — such as navigating to incorrect pages or triggering unintended actions — errors that current agents find particularly challenging to rectify. Our key insight is that we can leverage Molmo's general capabilities in conjunction with external feedback from system API to implement a self-validation mechanism, further enhancing the robustness of visual grounding to achieve near-human perfection.

Specifically, as shown in Figure~\ref{fig: grounding check}, when the grounding agent receives a click target description, it first attempts to generate the coordinates of the target in the screenshot. This attempt leads to one of two outcomes: either the agent determines that the target does not exist and outputs ``there are none", or it successfully outputs the coordinates and retrieves the corresponding element information using the function $\operatorname{get\_element\_info\_at\_position}(x, y)$ mentioned in Section~\ref{sec:action-recording}, including bounding box coordinates and element name. It then annotates the screenshot with red marks (similar to the marked screenshots used in cognition completion) indicating the element boundary and click position as a visual prompt, combining this with the retrieved element name to determine whether the generated position matches the initial target description. If it is judged to be inconsistent, the process reattempts output until the judgment passes or the retry limit is reached.

\section{Experiment}

\subsection{Experimental Setup}
To validate the effectiveness of our approach, we preliminarily conducted experiments on a relatively small scale, selecting the task of creating PowerPoint presentations as our primary testing ground. This choice is significant for mainly two reasons: First, presentation creation is a practical and common office task where automation can significantly reduce human workload. Second, it requires sophisticated cross-software coordination, particularly when creating high-quality presentations that involve web browsing to search and collect relevant materials. In future experiments, we’ll dive into a broader range of real-world scenarios where completing a work demands substantial human cognition, to further validate the scalability and effectiveness of our approach.

\paragraph{Data Collection}

\begin{wrapfigure}{r}{0.36\textwidth}
    \centering
    \includegraphics[width=1\linewidth]{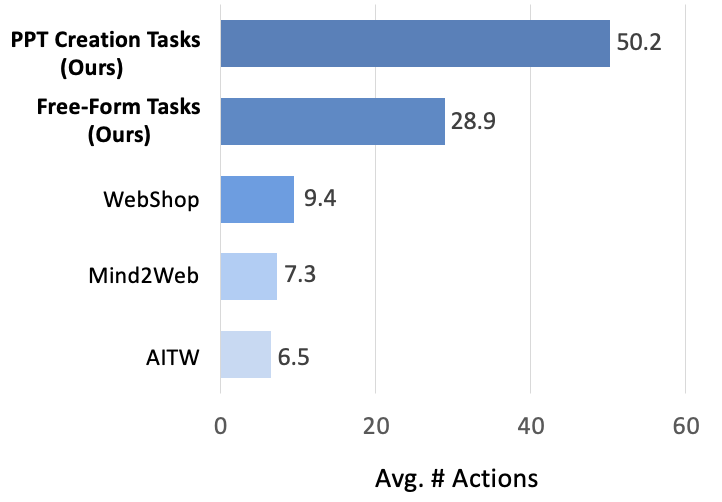}
    \caption{Average actions per task in our dataset vs. and other existing ones.}
    \label{fig: task number}
\end{wrapfigure}

Using PC Tracker's task oriented mode, we collected a dataset of 133 interaction trajectories, consisting of two categories: free-form Chrome and PowerPoint related tasks (30\%, 58 trajectories with an average of 29 events each) and guided PowerPoint presentation creation tasks (70\%, 75 trajectories with an average of 50 events each). It is worth noting that the average action number per task in our dataset is significantly larger than those in existing datasets, as shown in Figure~\ref{fig: task number}. The free-form trajectories include diverse but relatively simple operations such as web browsing, Chrome settings adjustment, and PowerPoint layout editing. In contrast, the guided presentation creation tasks follow specific presentation objectives, requiring users to engage in a more complex workflow that involves switching between PowerPoint and Chrome to collect online resources for presentation content, authentically simulating real-world work patterns.

\paragraph{Training}

We fine-tuned the Qwen2-VL-72B-Instruct model with our collected dataset. The training was conducted on 32 H100 GPUs for about 2 hours. We set the context length to 8,192 tokens.

\subsection{Evaluation}

\subsubsection{Why Not Existing Benchmarks}

While benchmarks like OSWorld \cite{xie2024osworldbenchmarkingmultimodalagents} play valuable roles in evaluating AI agents' basic capabilities of computer use, they are inappropriate for our specific evaluation needs. Existing benchmarks tend to focus on basic operations, like duplicating the last two slides in PowerPoint. However, creating high-quality presentations is significantly more complex. The deliverables of such tasks are highly variable, often requiring subjective assessments based on aesthetics, clarity, and alignment with human preferences, which largely extends beyond the scope of existing benchmarks. Furthermore, given that our agent is currently specialized in PowerPoint presentation creation, utilizing general benchmarks to assess its performance would be both insufficient and inappropriate.

\subsubsection{Human Evaluation and Case Studies}
\label{sec:human-eval}

Based on these considerations, we employed human evaluation to assess PC Agent's performance in real-world application scenarios. In our target domain — creating high-quality presentations — PC Agent exhibited outstanding performance. Its capability to execute dozens of steps and the quality of generated presentations substantially surpass existing solutions, as shown in Figure~\ref{fig: ppt example}. We examined two representative use cases:

\begin{figure}[h]
    \centering
    \includegraphics[width=1\linewidth]{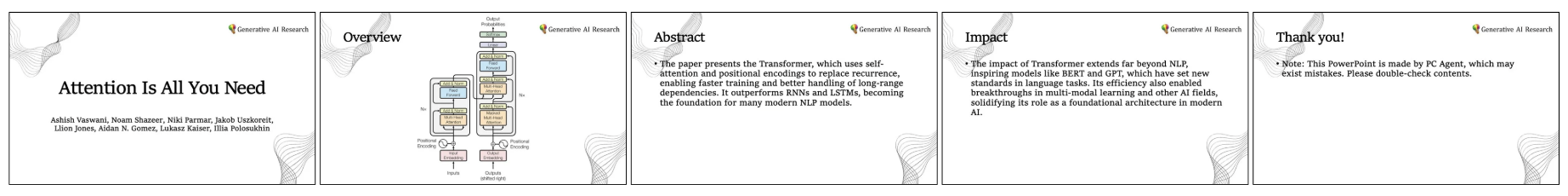}
    \caption{An example presentation created by PC Agent, see task description in Appendix~\ref{sec:task-description}.}
    \label{fig: ppt example}
\end{figure}


\paragraph{Case1: Complex Presentation Design}

We challenged PC Agent to create sophisticated presentations under detailed task specifications. It successfully accomplished the following objectives: 1) switching between PowerPoint and Chrome for resource collection, 2) organizing content into presentations with pictures, and 3) saving completed presentations to the desktop and closing all applications. Through human evaluation, we found that our agent successfully handles sequences of up to 50 steps to produce multi-page presentations with practical utility, showcasing its substantial potential for real-world applications.

\paragraph{Case2: Batch Processing}
We evaluated PC Agent's capability to handle repetitive tasks, such as creating multiple slides with consistent styling but varying content — a representative scenario where AI can significantly reduce human workload. Using a predetermined PowerPoint poster template, we tasked PC Agent with creating introduction posters for Turing Award laureates, a process requiring approximately 20 correct steps, including web-based image searches. Our assessment revealed significant potential in handling repetitive tasks: In an attempt to create 20 posters, PC Agent completed 11 posters meeting the requirements, achieving a 55\% success rate. It is important to note that in some failed cases, PC Agent did create a poster, but it had subtle deviations from human requirements. This is a situation that existing benchmarks find hard to assess.

\subsubsection{Analysis}

\paragraph{Failed Case Study}
Our analysis revealed that mistakes stem from various ways during execution. However, these errors share a common challenge: the agent's limited ability to recover once mistakes occur. Although our multi-agent workflow can prevent some planning mistakes through the grounding agent's feedback, incorrect plans that get executed often lead to unrecoverable failures, indicating the need for better error recovery capabilities.

\begin{wrapfigure}{r}{0.45\textwidth}
    \centering
    \includegraphics[width=1\linewidth]{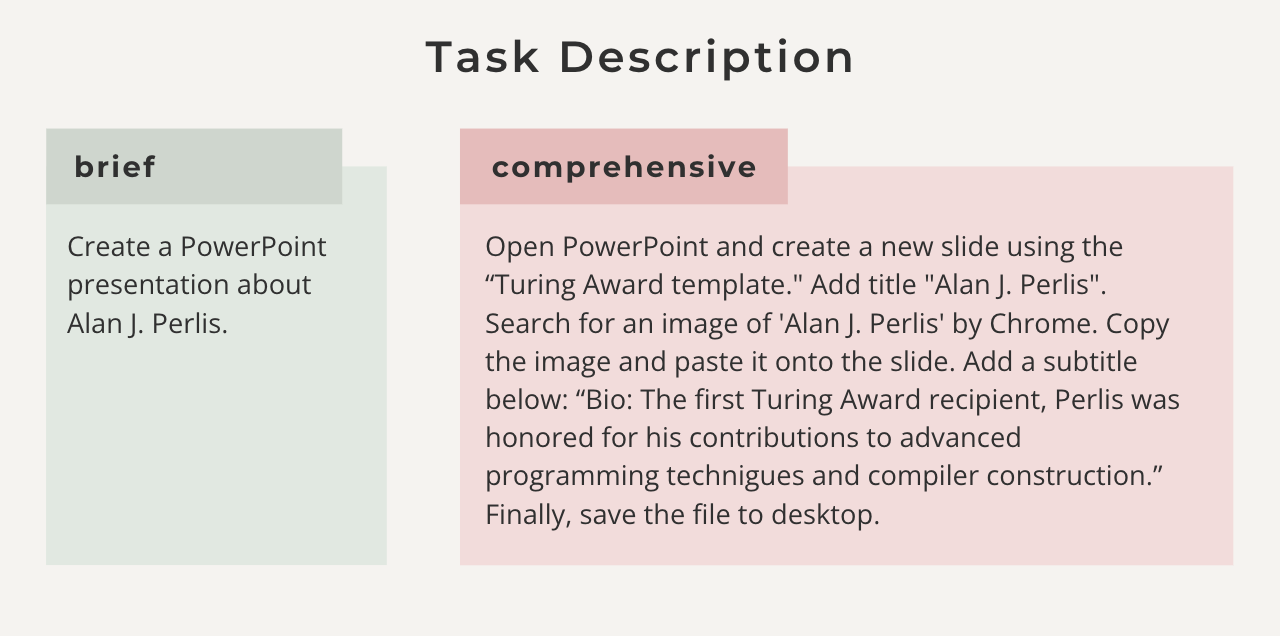}
    \caption{The brief and comprehensive task description for creating a poster about Alan J. Perlis.}
    \label{fig: task description}
\end{wrapfigure}

\paragraph{Impact of Task Description}
During our evaluation, we observed that the level of detail in task descriptions significantly influences PC Agent's performance, as exemplified in Figure~\ref{fig: task description}. Comprehensive task descriptions enable it to better understand human expectations, leading to more satisfactory outcomes. In contrast, brief task descriptions often result in insufficient understanding of requirements, causing the agent to favor quick but superficial solutions. Our findings suggest that optimizing task description quality is crucial for improving agent performance.

\paragraph{Generalization Capability}
Our evaluation suggests that PC Agent possesses adequate performance on web browsing tasks, despite the relatively limited proportion of Chrome-related training data. This observation suggests that the operational skills acquired in specific scenarios may contribute to improved adaptability across a broader range of applications.

\paragraph{Data Efficiency}
With a relatively small training dataset of just 133 trajectories, PC Agent demonstrates remarkable task processing capabilities, particularly in PowerPoint presentation creation tasks. This data efficiency primarily stems from the high quality of the human cognitive trajectories we constructed, demonstrating the effectiveness of our framework.

\section{Conclusion and Outlook}

In conclusion, we presented a cognition transfer framework that efficiently guides AI to the digital world through three key components: PC Tracker for collecting human-computer interaction data, a two-stage post-processing for cognition completion, and a multi-agent system for computer task automation.
Looking forward, we identify several potential directions:

\paragraph{Scaling and Generalization} While our approach demonstrates effectiveness in presentation creation with limited training data, validating its generalization ability requires large-scale experimentation across diverse software applications and task domains.

\paragraph{Robustness and Long-term Planning} Recent advances in LLM reasoning, exemplified by OpenAI o1~\cite{o1}, demonstrate the potential for robust operation over extended sequences. These developments suggest new approaches to long-term planning and error recovery in computer tasks.

\paragraph{Action Optimization}

Complex mouse actions, particularly \textbf{drag} and \textbf{scroll}, present notable challenges. Drag actions are especially difficult for visual grounding because both the start and end points are often abstract positions. For instance, in resizing a rectangle, the end point is an abstract position that requires an understanding of appropriate layout. Meanwhile, scroll actions are limited by our current design choice of omitting cursor movement tracking for simplification. These challenges highlight opportunities to enhance spatial understanding for abstract drag actions and reevaluate the trade-off between simplicity and completeness in cursor tracking.

\paragraph{Task-free Data Utilization} PC Tracker's non-task oriented mode enables large-scale collection of natural human-computer interaction trajectories. These trajectories contain rich operation patterns and behavioral preferences, which can be utilized as pre-training data to learn basic interaction patterns, as supervised learning examples by inferring user intentions after the fact, or as reference behaviors for training reward models in reinforcement learning.

\paragraph{Complex work Evaluation} While existing benchmarks primarily focus on basic tasks with deterministic outcomes, complex real-world work like presentation creation and video editing requires more comprehensive evaluation frameworks. These frameworks need to assess deliverables through multiple dimensions, including human preference alignment, aesthetic quality, and overall completeness.

\paragraph{User-Friendly Task Specification} While detailed task descriptions improve model performance, requiring users to provide extensive instructions can hinder system usability. Future work should explore methods to balance model requirements with user experience, such as inferring complete requirements from partial descriptions or enabling interactive clarification of task details.

\section*{Acknowledgments}

We would like to thank Yan Ma, Ethan Chern, Xuefeng Li, Mingyan Yang and Xingyang Li for their valuable feedback on different stages of this work.

\newpage
\bibliographystyle{acl_natbib}
\bibliography{main}

\newpage

\appendix

\newtcbox{\buttonbox}{
    nobeforeafter,
    tcbox raise base,
    boxrule=0.4pt,
    top=2pt,
    bottom=2pt,
    right=4pt,
    left=4pt,
    arc=3pt,
    boxsep=0pt,
    colback=white,
    colframe=black}


\section{PC Tracker User Manual}

\subsection*{1. Introduction}
PC Tracker is a lightweight infrastructure for efficiently collecting large-scale human-computer interaction trajectories. The program runs seamlessly in the background, automatically capturing screenshots and keyboard \& mouse activities. Figure~\ref{fig: trajectory example} shows an example of the collected human-computer interaction trajectories.  

\subsection*{2. Installation}
\begin{itemize}[itemsep=0pt]
    \item Ensure your operating system is Windows.
    \item Extract our software package to a location with sufficient disk space (recommended to have more than 3GB of available space for storing recorded data).
\end{itemize}

\subsection*{3. Quick Start}
\begin{itemize}[itemsep=0pt]
    \item (Optional) Set screen resolution to 16:9 (recommended 1920 x 1080).
    \item Open the extracted folder and launch \texttt{main.exe}.
\end{itemize}

\subsection*{4. Instructions}
After launching the tracker, you can choose between \textbf{Task Oriented Mode} or \textbf{Non-Task Oriented Mode} for recording.

\subsubsection*{Task Oriented Mode}
This mode is divided into two sub-modes: \textbf{Given Task} and \textbf{Free Task}.

\paragraph{Given Task}  
In this mode, you will be assigned an uncompleted task each time.
\begin{itemize}[itemsep=0pt]
    \item \textbf{Next Task}: Click \buttonbox{\textsf{Next Task}} to get the next task.
    \item \textbf{Previous Task}: Click \buttonbox{\textsf{Previous Task}} to return to the previous task.
    \item \textbf{Bad Task Feedback}: If you think the current task is difficult to complete, click \buttonbox{\textsf{Bad Task}} to discard it permanently. Alternatively, you can start the task and modify its description after completion based on your actual execution.
    \item \textbf{Start Recording}: Click \buttonbox{\textsf{Start}}, and the tracker window will automatically minimize while recording begins.
    \item \textbf{End Task}: After completing the task, click \buttonbox{\textsf{Finish}} to save the record. Or if the task execution fails or you don't want to record it, click \buttonbox{\textsf{Fail}}.
    \item \textbf{Modify Task Description}: After finishing the task, you can modify the task description based on your actual execution.
\end{itemize}

\paragraph{Free Task}  
In this mode, you can freely use the computer and summarize the task description and difficulty yourself.
\begin{itemize}[itemsep=0pt]
    \item \textbf{Start Recording}: Click \buttonbox{\textsf{Start}}, and the tracker window will automatically minimize while recording begins.
    \item \textbf{Save and Summarize This Record}: 
    Fill in the task description, select difficulty (easy/medium/hard), and click \buttonbox{\textsf{Save}} to save the record.
    \item \textbf{Discard This Record}: Click \buttonbox{\textsf{Discard}} to discard the record.
\end{itemize}

\subsubsection*{Non-Task Oriented Mode}

In this mode, you can freely use the computer, with similar methods to start and stop recording as described above.

\subsection*{5. Usage Notes}
\begin{itemize}[itemsep=0pt]
    \item \textbf{Does not currently support using extended screens.}
    \item \textbf{Does not currently support using Chinese input methods.}
    \item \textbf{Does not currently support using touchpads.}
    \item \textbf{The tracker window is fixed in fullscreen.} To support the filtering of tracker-related actions (such as clicking the Start button) in post-processing, the tracker window is fixed in fullscreen. You can reopen the tracker window by clicking to view the task description, then minimize it again, but please do not drag it to display in a non-fullscreen state.

\end{itemize}

\subsection*{6. Data Privacy}
\begin{itemize}[itemsep=0pt]
    \item After starting recording, your screenshots and keyboard \& mouse operations will be automatically recorded. PC Tracker does not record any information from unopened software. If you believe the recording may infringe on your privacy, you can choose to discard the record.
    \item Collected data is saved in the \texttt{./events} folder (hidden by default). Each trajectory includes a Markdown file for easy visualization.
\end{itemize}

\subsection*{7. FAQ}
\paragraph{Does the tracker have networking capabilities?}  
PC Tracker is completely local, does not support networking, and will not upload your data.

\paragraph{What if my computer screen resolution is not 16:9?}  
If your screen resolution is not 16:9, it will affect the subsequent unified processing of data. We recommend adjusting your screen resolution to 16:9.

\paragraph{How much space does the collected data occupy?}  
The specific data size varies. Generally, even with intensive recording operations for 1 hour, it will not generate more than 1GB of data.

\subsection*{8. Contact}
If you have any questions, please contact us at \href{mailto:henryhe_sjtu@sjtu.edu.cn}{henryhe\_sjtu@sjtu.edu.cn} or \href{mailto:zizi0123@sjtu.edu.cn}{zizi0123@sjtu.edu.cn}.

\newpage

\section{Prompts for Cognition Completion}
\subsection{Action Semantic Completion}
The first stage of cognition completion is action semantic completion. Specifically, we first generate descriptions for the click targets, and than refine the generated descriptions. Table~\ref{description gen} and Table~\ref{evaluation} present the prompts used for these two processes.
\lstset{
    basicstyle=\ttfamily\small,
    columns=flexible,
    breaklines=true,
    breakindent=0pt,
    literate={`}{{\textasciigrave}}1
             {√}{{\checkmark}}1
             {@}{{\texttimes}}1
}

\setlength{\parindent}{0pt}

\spaceskip=0.3em

\begin{table}[h!]
    \caption{Click Target Description Generation Prompt}
    \label{description gen}
    \renewcommand{\arraystretch}{1}
    \begin{tabular}{>{\setlength{\parskip}{0.18em}}p{0.95\textwidth}}
    \toprule
    \textbf{Click Target Description Generation Prompt} \\
    \midrule
    \begin{lstlisting}[basicstyle=\ttfamily\small]
    Help me describe the target in the screenshot. The target may be a GUI element or an empty area on the screen. 
    You will be provided with:
    1. A screenshot with a red mark quadruplet:
       - Frame: rectangular border around the target (may be inaccurate)
       - Circle: circle at the center of the target
       - Point: dot marking the exact click position
       - Arrow: pointing to the target
    2. The name of the clicked target for reference. It's just for reference. If this name is "Unknown" or appears to be incorrect, just ignore it.

    Description Rules:
    1. Priority Order:
       - Highest: Circle, Point and Arrow
       - Second: Reference name (if reliable)
       - Lowest: Frame

    2. Description Strategy:
       A. For Clear GUI Elements:
          - Include position info ("top", "left", "center", etc.) if possible
          - Use visual information to describe the element
          - Refer to the provided element name if reliable
          - Examples:
            √ "the button in the top-right corner of the window"
            √ "the current tab at the top of the browser"
            @ "the red circle" (red marks doesn't belong to the original screenshot or element)
       
       B. For Empty Areas or Uncertain Elements:
          - Focus on positional relationships
          - Use visual information to locate the position
          - Examples:
            √ "empty area on the right side of the window"
            √ "area near the bottom toolbar"
    
    3. Prohibited:
       - No speculation about element functions
       - No uncertain terms like "seems", "appears", "probably"
       - No description of elements outside the frame
    
    Output Format:
    - For GUI elements: "{position description} + {element description}"
    - For empty areas: "empty area + {position description}"

    Examples:
    √ "the close button in the top-right corner of the window"
    √ "the 'Chrome' icon on the desktop"
    √ "the left thumbnail panel in current window"
    √ "the 'Images' tab below the search bar"
    √ "'click to add title'"
    √ "the button in the top-right corner of the browser" (when the reference name is not reliable and you are unsure about the element)
    @ "what appears to be a settings button" (avoid speculation)
    \end{lstlisting}
    \end{tabular}
    \end{table}
    
    \clearpage
    \vspace*{-\topskip}
    \vspace{0pt}
    \begin{table}[!t]
    \setlength{\aboverulesep}{0pt}
    \setlength{\belowrulesep}{0pt}
    \setlength{\extrarowheight}{0pt}
    \begin{tabular}{>{\setlength{\parskip}{0.2em}}p{0.95\textwidth}}
    \begin{lstlisting}[basicstyle=\ttfamily\small]
    Important:

    1. Carefully observe the screenshot and the red mark quadruplet. Use these visual cues to describe the element or position as accurately as possible. But **DO NOT** explicitly state the red marks in your description. Avoid phrases like "red arrow marking on the slide.." or "the red circle.." 
    2. When uncertain, prefer positional description over semantic or functional speculation. Be extraordinarily cautious to avoid hallucinations.
    3. Be precise and output the description directly in an objective tone. Avoid sentences starting with "the target is","The pointed target is", or "it appears to be".
    
    4. Do not directly use the provided element name, create your own natural description based on visual information.

    Note:
    1. For the name of the clicked target for reference, it is either very precise or completely worthless. Judge its reliability based on visual information.
    If unreliable, ignore it and be cautious, preferably using only positional descriptions; if reliable, try to expand on its description as much as possible.

    2. Special cases: for the text box in PowerPoint, the name of the clicked target is usually "click to add title" or "click to add text".
    - "'click to add title'": for the title text box above the content text box or on the cover slide
    - "'click to add text'": for the content text box below the title text box
    - "'click to add subtitle'": for the subtitle text box below the title text box
    - "'the left thumbnail panel in current window'": for the **left slides thumbnail panel in PowerPoint**. But **DO NOT** abuse the use of "thumbnail" in other cases.
    \end{lstlisting} \\
    \bottomrule
    \end{tabular}
    \end{table}

\begin{table}[h!]
    \caption{Click Target Description Refinement Prompt}
    \label{evaluation}
    \renewcommand{\arraystretch}{1}
    \begin{tabular}{>{\setlength{\parskip}{0.1em}}p{0.95\textwidth}}
    \toprule
    \textbf{Click Target Description Refinement Prompt} \\
    \midrule
    \begin{lstlisting}[
        basicstyle=\ttfamily\footnotesize,  % 使用更小的字体
        basewidth={0.5em,0.5em},           % 减少字符间距
        lineskip=-0.5pt,                   % 减少代码行间距
        breaklines=true,
        breakindent=0pt
    ]
    You are provided with the following information about a mouse click on a computer screen:

    1. A screenshot showing:
       - A red dot and circle marking the exact click location
       - A red arrow pointing to the click location
       - A red box outlining the general area of the clicked element
       Note: While the dot, circle, and arrow are precise, the box might be less accurate

    2. The exact coordinates of the mouse click
   
    3. The element name from the accessibility tree
       Note: This information might be incomplete, with many elements labeled as "unknown".
   
    4. A pre-generated description of the click location
       Types:
       - Empty area description (e.g., "empty area near the bottom toolbar")
       - Specific element description (e.g., "the start button on the left corner of the taskbar")

    # Your Task
    Evaluate the provided description, determine if it is accurate. If not, provide the correct description. You can describe it as an empty area or a specific element. Do not mention the red marks on the screenshot.
    \end{lstlisting}
    \end{tabular}
    \end{table}
    \clearpage
    \vspace*{-\topskip}
    \vspace{0pt}
    \begin{table}[!t]
    \setlength{\aboverulesep}{0pt}
    \setlength{\belowrulesep}{0pt}
    \setlength{\extrarowheight}{0pt}
    \begin{tabular}{>{\setlength{\parskip}{0.1em}}p{0.95\textwidth}}
    \begin{lstlisting}[basicstyle=\ttfamily\small]
    # Critical Evaluation Points
    1. **Priority of Visual Evidence**: The red markers (dot, circle, arrow) on the screenshot show the ACTUAL click location. This is your primary source of truth.But **DO NOT** explicitly state the red marks in your description. Avoid phrases like "red arrow marking on the slide.." or "the red circle.." 
   
    2. **Element Name Usage**:
       - Ignore if marked as "unknown"
       - If available, use it to verify the description's accuracy
       - If there's a conflict between the element name and the description, carefully evaluate which is correct

    3. **Empty Area vs. Specific Element Distinction**:
       - True empty areas: Locations where clicks produce no effect
       - False empty areas: Locations that appear empty but are part of specific functional elements

    # Evaluation Process
    1. First, locate the exact click point using the red markers
    2. Check if the provided element name offers any useful information
    3. Determine if the location is a true empty area or part of a specific functional element
    4. Compare the given description against your findings
    5. Provide your response based on the required format

    # Important
    - Carefully determine the wrong description. Most of the time, the provided description is correct.
    - The pre-generated description may have hallucinations. Carefully evaluate it.

    Final Answer Format:(Response in English even the element name is Chinese)
    Thought Process: {your thought process}
    Answer:{your answer}

    Your answer should be either:
    - "Good" if the description is accurate
    - "Wrong. Correct Description: {your description}" if the description is inaccurate
    \end{lstlisting}\\
    \bottomrule
    \end{tabular}
    \end{table}

\newpage
\subsection{Thought Completion}
Based on the completed action semantics, we conduct thought process completion, prompt shown in Table~\ref{thought}.

\lstset{
    basicstyle=\ttfamily\small,
    columns=flexible,
    breaklines=true,
    breakindent=0pt,
    literate={`}{{\textasciigrave}}1
}

\setlength{\parindent}{0pt}
\spaceskip=0.3em

\begin{table}[h!]
    \caption{Thought Process Completion Prompt}
    \label{thought}
    \renewcommand{\arraystretch}{1}
    \begin{tabular}{>{\setlength{\parskip}{0.2em}}p{0.95\textwidth}}
    \toprule
    \textbf{Thought Process Completion Prompt} \\
    \midrule
    \begin{lstlisting}[basicstyle=\ttfamily\small]
    You are a helpful PC Agent designed to complete tasks on a computer. Your goal is to recreate your **thought process** behind a specific action.

    You will be provided with:

    1. The task you are attempting to complete.
    2. A history of the steps you have already performed (up to 50, if any; none if it was the first action).
    3. Subsequent actions (none if this is the last action).
    4. The specific action you chose to take.
    5. A screenshot of the computer screen at the moment you decided to take the action  
    6. The red marks on the screenshot:
    A. For Click Actions (click, right click, double click):
        - Frame: rectangular border around clicked element
        - Center: circle at element center
        - Click: point at exact click position
        - Arrow: pointing to clicked element
    B. For Drag Actions:
        - Start: red point and circle
        - End: red point and circle  
        - Arrow: from start to end position

    Explanation of actions:
    1. **click element: <{element description}>**: Click the element described by `{element description}`.
    2. **right click element: <{element description}>**: Right-click the element described by `{element description}`.
    3. **double click element: <{element description}>**: Double-click the element described by `{element description}`.
    4. **drag from (x1, y1) to (x2, y2)**: Drag the mouse from the position (x1, y1) to (x2, y2).
    5. **scroll (dx, dy)**: Scroll with offsets (dx for horizontal movement, dy for vertical movement).
    6. **press key: key_content**: Press the `key_content` on the keyboard.
    7. **hotkey (key1, key2)**: Press the combination of `key1` and `key2`.
    8. **type text: text_content**: Type the text `text_content` on the keyboard.
    9. **wait**: Pause briefly, usually for system responses or screen updates.
    10. **finish**: Indicate the task has been completed.
    11. **fail**: Indicate the task has failed.

    Further explanation of drag operation: drag from (x1, y1) to (x2, y2) is a combination of press the mouse at (x1, y1) and drag it to (x2, y2). It might has following purposes:
    1. Move/Translate - Moving an element from position (x1,y1) to (x2,y2)
    Common scenarios:
    - Dragging a file/folder to a new location
    - Moving a UI element (window, widget) to a different position
    - Moving elements (shapes, text boxes, images) in a PowerPoint slide
    - Adjusting slider controls or resizing elements
    - Reordering items in a list or menu

    2. Range Selection - Selecting content within a rectangular region defined by (x1,y1) and (x2,y2) as diagonal points
    Common scenarios:
    - Selecting multiple files/icons in a folder
    - Selecting text in a document. This is usually performed before copy/cut/delete/adjust text operation. After this action, the selected text will be highlighted.
    \end{lstlisting}
    \end{tabular}
    \end{table}
    
    \begin{table}[h!]
    \begin{tabular}{>{\setlength{\parskip}{0.2em}}p{0.95\textwidth}}
    \begin{lstlisting}[basicstyle=\ttfamily\small]
    - Selecting cells in a spreadsheet
    - Drawing selection rectangle on a canvas
    
    Consider the following to give your thought process:
    1. The current state of the screen and your last step (if any). Does current state align with your last plan? Are this action trying to fix something?
    2. Based on the history steps, how far have you progressed in the whole task? And based on your subsequent actions, what is the expected outcome of this action? (**DO NOT** explicitly state the next action in your output.)
    3. Based on all the information (task, observation, history, future), if this action seems not related to the task, is it possibly exploring the environment?
    Based on the above, recreate your thought process in a clear, natural first-person narrative.

    Other requirements:
    1. Be confident in your thought process. Avoid speculative or uncertain phrases like "it seems" or "this action might have been for."
    2. You may reference future actions as context, but **DO NOT** explicitly state the next action in your explanation.
    3. If there are red marks on the screenshot, you should use them to understand the action, but **DO NOT** explicitly state the red marks in your explanation. Avoid phrases like "I notice the red circles around..." or "the red arrow indicates...".
    3. Keep your explanations **concise and short**, do not conduct meaningless analysis and emphasis.
    4. Do not repeat the action after your thought process.

    Here are some examples of the thought process:
    - "I see the 'View' menu is successfully opened, so I can click the 'Slide Master' button on it to change the font for all slides."
    - "To open the target powerpoint file, I should open the folder containing it first. So I need to click the folder icon on the left of the taskbar."
    - "I want to click the close button to close the current window in my last step, but I see it is not closed yet in current screen. Maybe my click was not precise last time, so I need to click it again. I should click the close button on the right top corner of the window."
    - "After save the file to desktop, I have successfully completed the task."
    - "I need to modify the 5th slide, but it is not in the current screen. I should scroll down the page to find it."
    - "I have insert a new text box and focus on it, so I can type the content now."
    - "I have finished typing content in the text box. Now I can click anywhere outside the text box to deselect it and view the content on the slide."
    - "I see the current file name is 'Untitled', so I should change it to a proper name. First I need to click the text box of the file name to focus on it."
    - "I need to insert a new slide, so I can first click the left thumbnail panel in the PowerPoint window."
    - "I need to insert a new slide, and I have clicked the left thumbnail panel in the PowerPoint window. Now I need to press key enter to insert a new slide."

    Examples of thought processes for exploratory actions:
    - "I need to save the file to the desktop, but I don't see a desktop option in the window. Maybe I should scroll down to see if there's a desktop option."
    - "I want to select the save button, but I don't see a save option in the window. I guess I might find it by clicking the File button."
    - "I need to open the settings menu, but I don't see an obvious settings icon on the current interface. Perhaps I should click on the three dots or three horizontal lines icon in the top right corner, as these often hide more options."
    - "I want to change the document's font, but I can't find the font option on the toolbar. I might need to click on the 'Format' or 'Style' menu to see if I can find the font settings there."
    - "I need to insert an image, but I don't see an obvious 'Insert' button. I guess I might need to right-click on a blank area of the document to see if there's an option to insert an image in the context menu."
    - "I want to check the version information of this application, but I can't find the relevant option on the main interface. Maybe I should click on the 'Help' or 'About' menu, as version information is often found there."
    \end{lstlisting}
    \end{tabular}
    \end{table}
    
    \clearpage
    \vspace*{-\topskip}
    \vspace{0pt}
    \begin{table}[!t]
    \setlength{\aboverulesep}{0pt}
    \setlength{\belowrulesep}{0pt}
    \setlength{\extrarowheight}{0pt}
    \begin{tabular}{>{\setlength{\parskip}{0.2em}}p{0.95\textwidth}}
    \begin{lstlisting}[basicstyle=\ttfamily\small]
    - "I need to exit this full-screen program, but I don't see an exit button. I can try pressing the ESC key or moving the mouse to the top of the screen to see if a hidden menu bar appears."
    - "I want to search for specific content on this webpage, but I don't see a search box. I can try using the shortcut Ctrl+F (or Command+F) to see if it brings up the in-page search function."
    
    Additional PowerPoint Operation Tip:
    - These steps are to add a new slide at the end of the presentation:
    1. Click in the left thumbnail panel of the PowerPoint window.
    2. Press the Enter key to insert a new slide.
    - These steps are to add text in the text box:
    1. Click 'click to add text'/'click to add title'/'click to add subtitle' to focus on the text box.
    2. Type the content in the text box.
    3. (Optional) Press the Enter key to finish.

    Again, you are recreating your thought process when you made the action, so you should not include any post-event evaluation or similar phrases.
    
\end{lstlisting} \\
\bottomrule
\end{tabular}
\end{table}

\newpage
\section{Task Description Example}
\label{sec:task-description}

\begin{table}[h!]
    \caption{The task description for Figure~\ref{fig: ppt example}}
    \setlength{\parskip}{0pt}
    \setlength{\baselineskip}{0.8em}
    \renewcommand{\arraystretch}{0.8}
    \begin{tabular}{>{\setlength{\parskip}{0.1em}}p{0.95\textwidth}}
    \toprule
    \textbf{Task Description Example} \\
    \midrule
    \begin{lstlisting}[
        basicstyle=\ttfamily\footnotesize,
        basewidth={0.5em,0.5em},
        lineskip=-0.5pt,
        breaklines=true,
        breakindent=0pt
    ]
    Open PowerPoint and create a new presentation using the "white template". 

    For the first slide, set the title to "Attention Is All You Need" and the subtitle to "Ashish Vaswani, Noam Shazeer, Niki Parmar, Jakob Uszkoreit, Llion Jones, Aidan N. Gomez, Lukasz Kaiser, Illia Polosukhin." 

    For the second slide, set the title to "Overview". Search for an image of "Transformer Attention Is All You Need" by Chrome and copy the source image. Back to PowerPoint and paste the image into the "Overview" slide. 

    For the third slide, title it "Abstract", with the following text: "The paper presents the Transformer, which uses self-attention and positional encodings to replace recurrence, enabling faster training and better handling of long-range dependencies. It outperforms RNNs and LSTMs, becoming the foundation for many modern NLP models." 

    For the fourth slide, title it "Impact", with the following text: "The impact of Transformer extends far beyond NLP, inspiring models like BERT and GPT, which have set new standards in language tasks. Its efficiency also enabled breakthroughs in multi-modal learning and other AI fields, solidifying its role as a foundational architecture in modern AI."

    For the last slide, add title "Thank you!" and following text: "Note: This PowerPoint is made by PC Agent, which may exist mistakes. Please double-check contents." 
    
    Save the presentation and close PowerPoint and Chrome window and you're all set!
    \end{lstlisting} \\
    \bottomrule
    \end{tabular}
    \end{table}

\end{document}